\documentclass[journal]{IEEEtran}
\usepackage{ dsfont }
\pdfoutput=1
\usepackage[T1]{fontenc}% optional T1 font encoding
\usepackage{cite}
\usepackage{amsmath,amssymb,amsfonts}
\usepackage{algorithmic}
\usepackage{graphicx}
\usepackage{epstopdf}
\usepackage{textcomp}
\usepackage{algorithm}
\usepackage{algorithmic}
\usepackage{subfigure}
\usepackage{caption}
\usepackage{color}
\usepackage{verbatim}
\usepackage{array}
\usepackage{booktabs} %调整表格线与上下内容的间隔
\usepackage{multirow}
\usepackage[backref]{hyperref}
\begin{document}
\title{Multi-View Adaptive Fusion Network \\ for 3D Object Detection}
%作者信息已添加了李清泉老师为通讯作者
%基金信息已补全且替换了顺序
\author{Guojun Wan
g$^\dagger$ , Bin Tian$^\dagger$, Yachen Zhang, Long Chen, Dongpu Cao, Jian Wu% <-this % stops a space
\thanks{This work was supported by the Key-Area Research and Development Program of Guangdong Province (2020B090921003), the National Natural Science Foundation of China (61503380, 61773381), and the Intel Collaborative Research Institute for Intelligent and Automated Connected Vehicles ("ICRI-IACV").}
\thanks{$^\dagger$G. Wang and B. Tian contributed equally to this work.}
\thanks{G. Wang is with the State Key Laboratory of Automotive Simulation and Control, Jilin University, Changchun 130022, China (email: 839977837wgj@gmail.com).}% <-this % stops a space
\thanks{B. Tian is with the State Key Laboratory of Management and Control for Complex Systems, Institute of Automation, Chinese Academy of Sciences, and with School of Artificial Intelligence, University of Chinese Academy of Sciences, Beijing 100190, China (e-mail: bin.tian@ia.ac.cn)}%
\thanks{Y. Zhang and L. Chen are with School of Data and Computer Science, Sun Yat-Sen University, 510275, Guangzhou, China. (e-mail: zhyachen@mail2.sysu.edu.cn and chenl46@mail.sysu.edu.cn)}
\thanks{D. Cao is with the Waterloo Cognitive Autonomous Driving (CogDrive) Lab, University of Waterloo, Canada. (e-mail: dongpu.cao@uwaterloo.ca)}
\thanks{Jian Wu (Corresponding author of this paper) is a Professor of College of Automotive Engineering at Jilin University. (e-mail: wujian@jlu.edu.cn)}}
\maketitle

\begin{abstract}
3D object detection based on LiDAR-camera fusion is becoming an emerging research theme for autonomous driving.
However, it has been surprisingly difficult to effectively fuse both modalities without information loss and interference.
To solve this issue, we propose a single-stage multi-view fusion framework that takes LiDAR bird's-eye view, LiDAR range view and camera view images as inputs for 3D object detection.
To effectively fuse multi-view features, we propose an attentive pointwise fusion (APF) module to estimate the importance of the three sources with attention mechanisms that can achieve adaptive fusion of multi-view features in a pointwise manner.
Furthermore, an attentive pointwise weighting (APW) module is designed to help the network learn structure information and point feature importance with two extra tasks, namely, foreground classification and center regression, and the predicted foreground probability is used to reweight the point features.
We design an end-to-end learnable network named MVAF-Net to integrate these two components.
Our evaluations conducted on the KITTI 3D object detection datasets demonstrate that the proposed APF and APW modules offer significant performance gains. Moreover, the proposed MVAF-Net achieves the best performance among all single-stage fusion methods and outperforms most two-stage fusion methods, achieving the best trade-off between speed and accuracy on the KITTI benchmark.
\end{abstract}

\begin{IEEEkeywords}
3D Object Detection, Multi-View, LiDAR-camera Fusion, Point Cloud.
\end{IEEEkeywords}

\IEEEpeerreviewmaketitle

\section{Introduction}
\IEEEPARstart{3}{D} object detection is particularly useful for autonomous driving applications because diverse types of dynamic objects, such as surrounding vehicles, pedestrians, and cyclists, must be identified in driving environments. Significant progress has been witnessed in the 3D object detection task with different types of sensors in the past few years, such as monocular images \cite{chen2016monocular}\cite{xu2018multi}, stereo cameras \cite{chen20173d}, and LiDAR point clouds\cite{luo2018fast}\cite{yang2018pixor}\cite{zhou2018voxelnet}. Camera images usually contain rich features (e.g., color, texture) while suffering from the lack of depth information and affected by light and weather. LiDAR point clouds provide accurate depth and geometric structure information, which are quite helpful for obtaining the 3D pose of objects. Thus far, various 3D object detectors employing LiDAR sensors have been proposed, including PIXOR \cite{yang2018pixor}, VoxelNet\cite{zhou2018voxelnet}, PointRCNN\cite{shi2019pointrcnn}, STD\cite{yang2019std}, SECOND\cite{yan2018second}, PointPillars\cite{lang2019pointpillars}, Part-A2\cite{shi2020points} and CenterNet3D\cite{wang2020centernet3d}. Although the performance of 3D object detectors that are only based on LiDAR have been significantly improved lately, most of them operate in a single-projection view, such as the bird's eye view (BEV) or range view (RV). BEV encoding of LiDAR points can preserve their physical dimensions, and objects are naturally separable. When represented in this view, however, point clouds are sparse and have highly variable point density, which may cause detector difficulties in detecting distant or small objects. On the other hand, RV encoding provides dense observations that allow more favorable feature encoding for such cases. The RV representation has been shown to perform well at longer ranges where the point cloud becomes very sparse, and especially on small objects. However, regardless of the chosen encoding of LiDAR returns, LiDAR point clouds are still limited in their capacity to provide color and texture information. Compared with LiDAR, RGB-images from camera view (CV) have much richer information to distinguish vehicles and background. However, they are sensitive to illumination and occlusions. Thus, the representations in different views have their own shortcomings, and only a single-view input is not sufficient for 3D object detection for autonomous driving. This motivates our intention to design an effective framework to fuse features from different views for accurate 3D object detection.
\begin{figure*}[tb]
  \centering
    \includegraphics[scale=0.123]{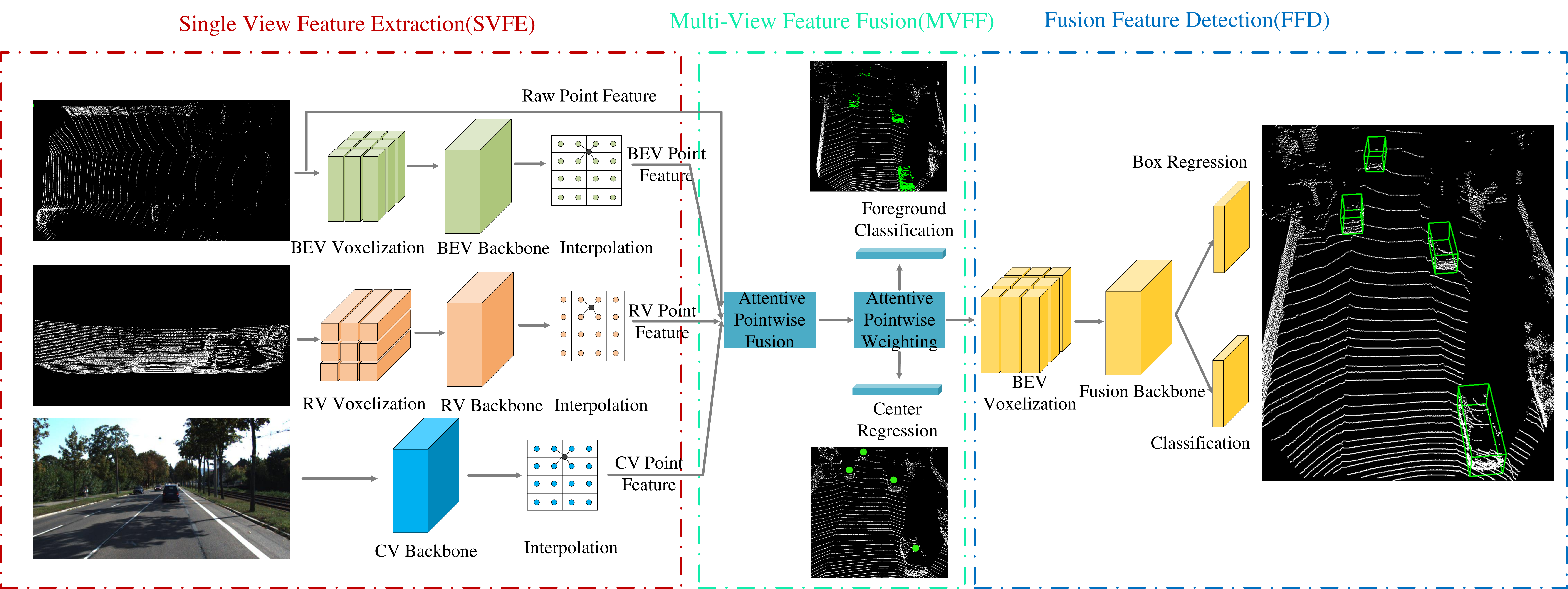}
    \captionsetup{font={footnotesize}}
    \caption{Overall architecture of MVAF-Net. The overall MVAF-Net consists of three parts: 1) single view feature extraction (SVFE), 2) multi-view feature fusion (MVFF) and 3) fusion feature detection (FFD). In the SVFE part, the raw RGB images and point clouds are processed by a three-stream CNN backbone (CV, BEV and RV backbone) to generate multi-view feature maps, where the point clouds are voxelized in both BEV and RV. In the MVFF part, the multi-view features are adaptively fused with the proposed attentive pointwise fusion module in a pointwise manner. The fused point features are further processed with the proposed attentive pointwise weighting module to reweight the point features and learn structure information. In the FFD part, the fused and reweighted point features are voxelized again and used as input to the fusion backbone for final 3D detection.}
    \label{fig1}
\end{figure*}
In fact, the problem of fusing features from different views is challenging, especially for LiDAR points and RGB images, as the features obtained from RGB images and LiDAR points are represented in different perspectives. When the features from RGB images are projected onto 3D LiDAR coordinates, some useful spatial information about the objects might be lost since this transformation is a one to-many mapping. Furthermore, the occlusions and illumination in RGB images may also introduce interference information that is harmful to the object detection task. Indeed, it has been difficult for the LiDAR-camera fusion-based methods to surpass the LiDAR-only methods in terms of performance. Furthermore, the large scale variations and occlusions in RV also introduce noise in the feature fusion process. To solve the above issues, various LiDAR-camera fusion strategies have been proposed for 3D object detection. The previous fusion strategies can be summarized as two main categories. 1) Result-level fusion. In these methods, information aggregation occurs at the result-level. The intuition behind these methods is to use off-the-shelf 2D object detectors to condense the regions of interest for the 3D object detectors\cite{qi2018frustum}\cite{wang2019frustum}\cite{xu2018pointfusion}. 2) Feature-level fusion. These methods jointly reason over multi-view inputs, the intermediate features of which are deeply fused\cite{chen2017multi}\cite{ku2018joint}\cite{liang2019multi}\cite{liang2018deep}\cite{zhu2020cross}\cite{sindagi2019mvx}\cite{vora2020pointpainting}\cite{zhou2020end}. Although effective, these methods have several limitations. Result-level fusion cannot leverage the complementarity among different views, and their performance is bounded by each stage. Furthermore, most of the previous feature-level fusion methods have been fusing features via simple concatenation or element-wise summation/mean operation, such as MV3D\cite{chen2017multi}, AVOD\cite{ku2018joint}, MMF\cite{liang2019multi}, MVX-Net\cite{sindagi2019mvx}, PointPainting\cite{vora2020pointpainting} and MVF\cite{zhou2020end}, they do not consider the mutual interference and importance of multi-view features. However, the RGB images from CV and range images from RV often have noisy information such as occlusion and truncation. Thus, the wrong point features will be obtained after 3D points are projected onto the RGB images or range images. Therefore, simply using pointwise projected features will degrade the performance of 3D detection. Recently, with the application of attention mechanisms in visual CNN models, some methods have adopted the self-attention mechanism to achieve multi-view feature fusion, such as PI-RCNN\cite{xie2020pi}, 3D-CVF\cite{yoo20203d} and EPNet\cite{huang2020epnet}. Although good performance has been achieved, these fusion methods generally adopt a two-stage technique that uses ROI-pooling to fuse the image with the point cloud. In addition, the heavy point cloud feature extractors, such as PointNet++\cite{qi2017pointnet2} and 3D Convolutions (Convs)\cite{yan2018second}, further increase the computational burden of the entire network.

In this paper, to address the above challenges, we design a multi-view adaptive fusion network named MVAF-Net that can fuse the spatial feature maps separately extracted from BEV, RV and CV effectively in a pointwise manner. It is an end-to-end single-stage LiDAR-image fusion 3D detection method without extra refinement stage and is only composed of efficient 2D Convs. The entire network consists of three parts, namely, single view feature extraction (SVFE), multi-view feature fusion (MVFF) and fusion feature detection (FFD), as shown in Fig. \ref{fig1}. In the SVFE part: the raw RGB images and point clouds are processed by three-stream feature extraction network to generate multi-view feature maps, where the point clouds are voxelized into pillars in both BEV and RV. In the MVFF part: the BEV, RV and CV point features are obtained by projecting the raw point clouds onto their respective feature maps and bilinear interpolation. To solve the challenges in multi-view feature fusion, we design an attentive pointwise fusion (APF) module to adaptively fuse the multi-view features in a pointwise manner. First, the BEV, RV and CV point features are concatenated to obtain preliminary mixed features; then, the mixed point features are used as the inputs of the APF module to estimate the channel-wise importance that determines how much information is brought from three sources using attention mechanisms. In this way, useful multi-view features are utilized to obtain the fused point features, while noisy features are suppressed. Compared with the previous methods, our solution possesses two main advantages, including 1) achieving fine-grained pointwise correspondence between multi-view inputs through pointwise projection; 2) addressing the issue of the interference information that may be introduced by CV and RV. To compensate for the loss of geometric structure information of point clouds in the voxelization process, we further enrich the fused point features with raw point features.

Intuitively, points belonging to the foreground objects should contribute more to final detection tasks, while those from the background regions should contribute less. Hence, we proposed an attentive pointwise weighting (APW) module  to reweight the fused point features with extra supervision from foreground classification. Inspired by SA-SSD\cite{he2020structure}, we add another supervision with center regression. Thus, the APW module uses the fused point features to perform two extra tasks: foreground classification to reweight the fused point features and center regression to instill an awareness of intra-object relationships in the features. Unlike SA-SSD, our APW module will operate during training and testing phases and is not auxiliary. In addition, our APW module uses point features instead of voxel center features, which can better maintain geometric structure information. In the FFD part, the fused and reweighted point features are voxelized again and used as inputs to the fusion backbone for final 3D detection. The contributions of our work are summarized as follows:

\begin{itemize}

\item We design an attentive pointwise fusion module that achieves fine-grained pointwise correspondence between multiple views through pointwise projection. The APF module can adaptively fuse features from BEV, RV and CV with attention mechanisms in a pointwise manner.
\item We propose an attentive pointwise weighting module with extra supervisions from foreground classification and center regression. The predicted foreground classification can be used to reweight the fused point features, and center regression can enforce learning structure information by the CNN backbone and achieve better localization performance without extra cost.
\item Based on attentive pointwise fusion and attentive pointwise weighting module, we propose a multi-view adaptive fusion network (MVAF-Net), a new 3D object fusion detection framework, that effectively combines information from multiple views: BEV, RV and CV in single stage. And it achieves the best performance among all single-stage fusion methods and outperforms most two-stage fusion methods on the KITTI benchmark.
\end{itemize}

\section{Related Work}

\subsection{LiDAR-only 3D Object Detection}
Currently, there are four types of point cloud representations used as input for 3D detectors. 1) Point-based representation\cite{shi2019pointrcnn}\cite{yang20203dssd}. The raw point cloud is directly processed, and bounding boxes are predicted based on each point. 2) Voxel-based representation \cite{yang2018pixor}\cite{zhou2018voxelnet}\cite{yan2018second}\cite{lang2019pointpillars}\cite{shi2020points}\cite{wang2020centernet3d}. The raw point clouds are converted to compact representations with 2D/3D voxelization. 3) Mixture of representations\cite{shi2020pv}\cite{he2020structure}. In these methods, both points and voxels are used as inputs, and their features are fused at different stages of the networks for bounding box prediction. The voxel-based methods discretize 3D point clouds into a regular feature representation, then a 2D/3D CNN backbone is applied for 3D bounding box prediction, such as VoxelNet\cite{zhou2018voxelnet}, SECOND\cite{yan2018second}, PointPillars\cite{lang2019pointpillars} and CenterNet3D\cite{wang2020centernet3d}. The voxel-based methods have high computational efficiency, however, the information loss is caused during the point cloud voxelization process. Point-based methods directly process raw LiDAR points using PointNet/PointNet++\cite{qi2017pointnet2} to yield the global feature representing the geometric structure of the entire point set, such as PointRCNN\cite{shi2019pointrcnn}, STD\cite{yang2019std} and 3DSSD\cite{yang20203dssd}. Compared to the voxel-based methods, the point-based methods have flexible receptive fields for point cloud feature learning with set abstraction operation, however, they are limited by high computation costs. The methods based on a mixture of representations take both point and voxel inputs and fuse their features at different stages of the networks for 3D object detection, such as PV-RCNN\cite{shi2020pv} and SA-SSD\cite{he2020structure}. These methods can take advantage of both the voxel-based operations (i.e., 3D sparse convolution) and PointNet-based operations (i.e., set abstraction operation) to enable high computational efficiency and flexible receptive fields for improving the 3D detection performance.

\subsection{Multi-modal Fusion based 3D Object Detection}

To take advantage of camera and LiDAR sensors, various fusion methods have also been proposed. According to the stages of fusion occurring in the whole detection pipeline, they can be summarized into two main categories, including result-level fusion and feature-level fusion. The result-level fusion methods leverage image object detectors to generate 2D region proposals to condense regions of interest for the 3D object detectors\cite{qi2018frustum}\cite{wang2019frustum}\cite{xu2018pointfusion}. However, the performance of these methods is limited by the accuracy of the camera-based detectors. The feature-level fusion methods jointly reason over multi-sensor inputs and the intermediate features of which are deeply fused \cite{chen2017multi}\cite{ku2018joint}\cite{liang2019multi}\cite{liang2018deep}\cite{zhu2020cross}\cite{sindagi2019mvx}\cite{vora2020pointpainting}. MV3D\cite{chen2017multi} is a pioneering work of this type of method, which takes CV, RV, and BEV as input, and exploits a 3D RPN to generate 3D proposals. AVOD\cite{ku2018joint} fused the LiDAR BEV and CV features at the intermediate convolutional layer to propose 3D bounding boxes. ContFuse\cite{liang2018deep} uses continuous convolution to fuse images and LiDAR features on different resolutions. MMF\cite{liang2019multi} adds ground estimation and depth estimation to the fusion framework and learns better fusion feature representations while jointly learning multi-tasks.

While various sensor fusion networks have been proposed, they do not easily outperform LiDAR-only detectors because they have seldom recognized different aspects of the importance and noise of multi-view features. In the next sections, we will present our proposed MVAF-Net to overcome this challenge.

\section{Multi-View Adaptive Fusion Network}

\begin{figure*}[tb]
    \centering
    \includegraphics[scale=0.187]{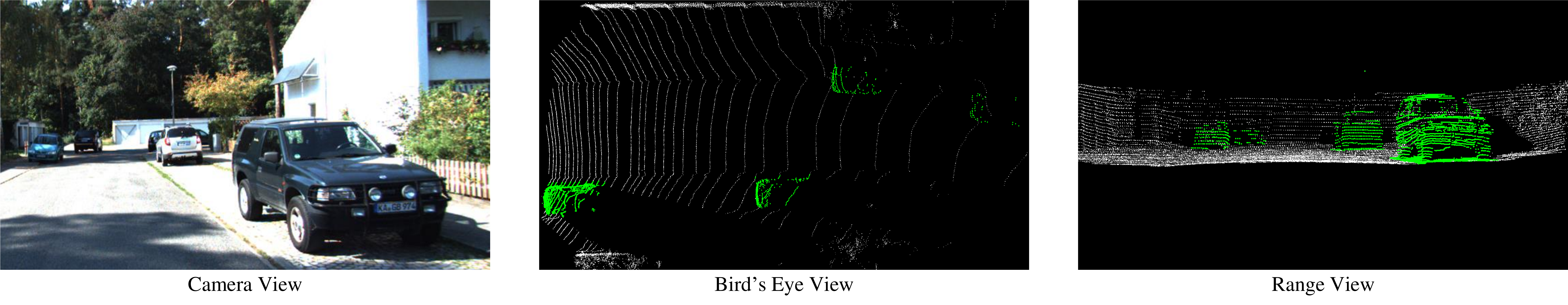}
    \captionsetup{font={footnotesize}}
    \caption{Multi-view inputs of MVAF-Net.}
    \label{fig2}
\end{figure*}

The overall architecture of the proposed MVAF-Net is illustrated in Fig.~\ref{fig1}. It is an end-to-end single-stage fusion 3D detection method consisting of three parts: single view feature extraction (SVFE), multi-view feature fusion (MVFF) and fusion feature detection (FFD). In the SVFE part: the raw RGB images and point clouds are processed by the three-stream CNN backbone (CV, BEV and RV backbone) to generate multi-view feature maps, where the point clouds are voxelized both in BEV and RV. In the MVFF part: the multi-view features are adaptively fused with the proposed attentive pointwise fusion module in a pointwise manner. The APF module can adaptively determine how much information is introduced from the three sources using attention mechanisms. The fused point features are further processed with the proposed attentive pointwise weighting module to reweight the point features and learn structure information. In the FFD part, the fused and reweighted point features are voxelized again and used as input to the fusion backbone for final 3D detection.

For simplicity in this paper, the BEV, RV and fusion backbone all use the same network architecture but different numbers of blocks. They all have two sub-networks: one top-down network that produces features at increasingly small spatial resolution and a second network that performs upsampling and concatenation of the top-down features. The top-down backbone can be characterized by a series of convolution blocks. Each block has several 3x3 2D Conv layers followed by BatchNorm and Leaky ReLU. The first Conv layer inside each block has a stride of 2 to downsample the feature maps. The final features from each top-down block are upsampled to the same size using transposed 2D Convs. Finally, we combine them in a concatenation manner and obtain a more representative feature map containing rich and semantic image information with different receptive fields. The CV backbone can be any pretrained image classification network for feature extraction of RGB images.

\subsection{Single View Feature Extraction}
\subsubsection{Camera View Stream}

The image stream takes RGB images as input and extracts the semantic information with the CV backbone. In this paper, we choose the lightweight RegNetX\cite{radosavovic2020designing} as the CV backbone because of its efficiency. The CV backbone has four blocks to downsample inputs by 16$\times$, and the outputs of the last three blocks are upsampled and concatenated to obtain CV feature maps $F_{CV}$.

\subsubsection{Bird's-Eye View Stream}

 Given that BEV encoding can maintain the physical dimension and scale information, as shown in Fig.~\ref{fig2}, we voxelize the point cloud into pillars in BEV. To reduce information loss and improve memory efficiency, we use dynamic voxelization to discretize the point cloud. The dynamic voxelization is proposed by\cite{zhou2020end}. It can avoid the unstable performance caused by non-deterministic voxel embedding and unnecessary memory usage caused by voxel padding.

Specifically, dynamic voxelization in this paper consists of three steps: 1) Crop the point clouds based on the ground-truth distribution with range $L$, $W$, $H$ along the $x$, $y$, $z$ axes, respectively. A point-based fully connected layer is adapted to learn high-dimensional point features. 2) The high-dimensional point features are then grouped into pillars with a voxel size of $v_x$, $v_y$ along $x$, $y$ axes and a max pooling operation is employed to obtain pillar features. 3) The encoded pillar features are scattered back to the original pillar positions to construct a pseudo-image that is further processed by the BEV backbone. The BEV backbone has three blocks to downsample  the pseudo-image by 8$\times$, and the outputs of the three blocks are upsampled and concatenated to obtain the BEV feature maps $F_{BEV}$.

\subsubsection{Range View Stream}

RV encoding is the native representation of the rotating LiDAR sensor. It retains all original information without any loss. Beyond this, the dense and compact properties make it efficient to process. Thus, we propose another range view stream to extract point cloud features.

Analogous to\cite{wang2020pillar}, we discretize the point clouds in the cylindrical coordinate system. Compared with the spherical coordinate system used in\cite{zhou2020end}, the cylindrical coordinate system can better maintain the scale in the z-axis direction. The cylindrical coordinates ($\rho_i$,$\phi_i$,$z_i$) of a point $p_i$ ($x_i$,$y_i$,$z_i$) are given by the following:
\begin{equation}
\rho_i = \sqrt{x^2 + y^2}, \quad
\phi_i = \arctan \frac{y_i}{x_i}, \quad
z_i = z_i
\label{eq1}
\end{equation}

The same dynamic voxelization operation is used for feature extraction in the range view. First, a point-based fully connected layer is adapted to learn high-dimensional point features based on the cylindrical coordinates and intensity of points. The high-dimensional cylindrical point features are also grouped into pillars with a voxel size of $v_\phi$, $v_z$ along $\phi, z$ axes and a max pooling operation is employed to obtain cylindrical pillar features. The encoded cylindrical pillar features are also scattered back to the original positions to construct a cylindrical pseudo-image feature map that is further processed by RV backbone. The RV backbone also has three blocks to downsample  the pseudo-image 8$\times$, and the outputs of the three blocks are upsampled and concatenated to obtain the RV feature maps $F_{RV}$.

\subsection{Multi-View Feature Fusion}

In the MVFF part, we perform multi-view feature fusion and learning in a pointwise manner with our proposed modules: the APF module and APW module. The APF module adaptively fuses multi-view features with attention mechanisms which can determine how much information is introduced from the three sources. The fused point features are further processed with the proposed APW module to reweight the fused point features and learn structure information.

\subsubsection{Multi-View Feature Mapping}
 \begin{figure}[tb]
  \centering
    \includegraphics[scale=0.3]{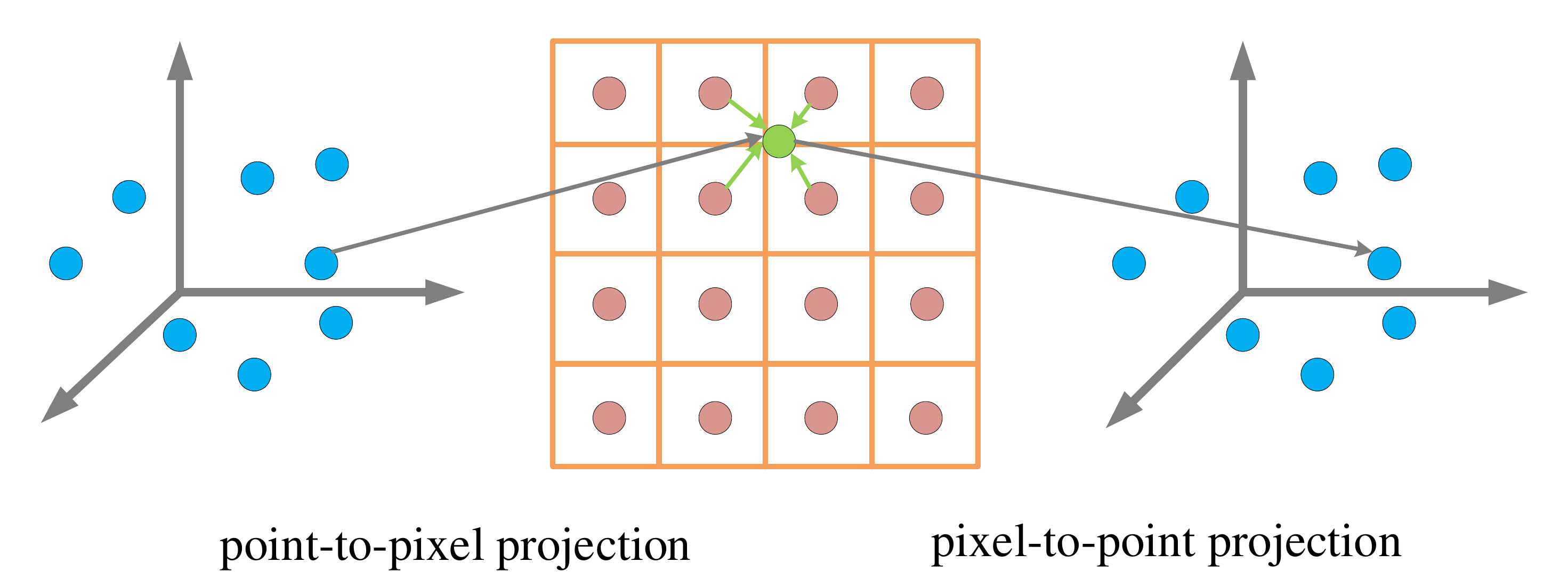}
    \captionsetup{font={footnotesize}}
    \caption{Illustration of multi-view pointwise feature mapping using bilinear interpolation.}
    \label{fig3}
\end{figure}
To fuse multi-view features in a pointwise manner, we should establish the correspondence between raw point clouds and the multi-view feature maps from the above three-stream backbone. Concretely, we project the LiDAR points onto $F_{CV}$, $F_{RV}$, $F_{BEV}$, using mapping matrixes $M_{CV}$, $M_{RV}$, $M_{BEV}$, respectively. In more detail, for a particular point $p_i$($x_i$, $y_i$, $z_i$) in a point cloud, we can obtain its corresponding position $p_{i}^{\prime}$($x_{i}^{\prime}$, $y_{i}^{\prime}$) in the $F_{CV}$, which can be written as:
\begin{equation}
p_{i}^{\prime} = M_{CV} * p_i
\label{eq2}
\end{equation}
Similarly, we can obtain the corresponding positions in the $F_{RV}$ and $F_{BEV}$. Where, $M_{CV}$ is the LiDAR-camera-view projection matrix, and $M_{RV}$ and $M_{BEV}$ can be obtained from the voxelization parameters in RV stream and BEV stream, respectively. Then, the multi-view point features can be obtained by bilinear interpolation on the corresponding position, as shown in Fig.\ref{fig3}.

\subsubsection{Attentive Pointwise Fusion}
\begin{figure}[tb]
    \centering
    \includegraphics[scale=0.34]{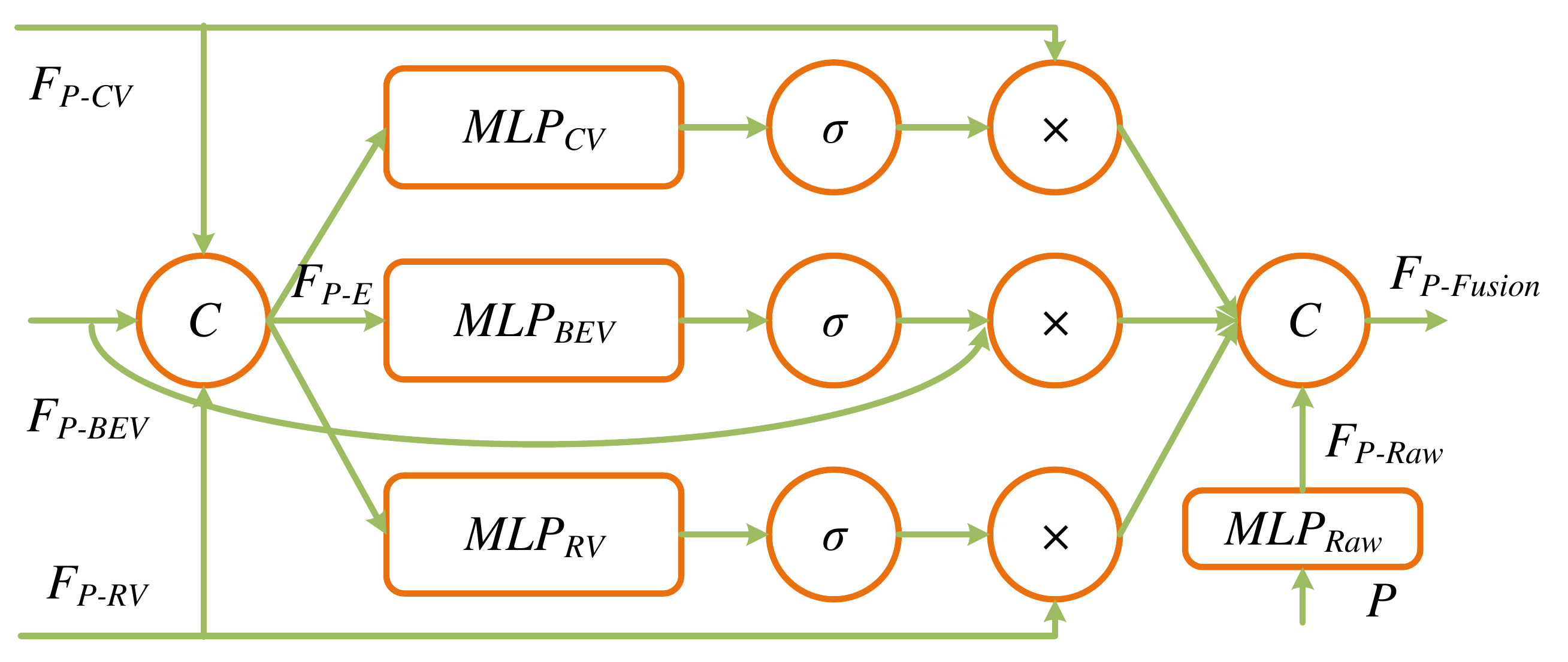}
    \captionsetup{font={footnotesize}}
    \caption{Architecture of the proposed attentive pointwise fusion module.}
    \label{fig4}
\end{figure}
To extract only essential features from multi-view point features, we propose an attentive pointwise fusion module that selectively combines multi-view point features depending on the relevance to the object detection task. The proposed APF module is depicted in Fig.\ref{fig4}. The multi-view point features $F_{P-CV}$, $F_{P-BEV}$ and $F_{P-RV}$ are concatenated channel-wise to obtain extended point features $F_{P-E}$. The extended features $F_{P-E}$ are then fed into three channel-wise attention modules separately, each of which uses the extended features to adaptively estimate their respective importance in a channel-wise manner. Specifically, the extended features are fed into their respective fully connected layers ${MLP}_{BEV}$, ${MLP}_{CV}$ and ${MLP}_{RV}$, each of which includes a linear layer, a ReLU layer, and a linear layer. Then, the respective feature weights are obtained through a sigmoid function, and finally the respective attention features are obtained by multiplying the weights with the corresponding features in a channel-wise manner. The specific forms of channel attention are as follows:
%%补充公式：
\begin{equation}
\begin{aligned}
\left\lbrace
\begin{array}{l}
F_{a.P-CV} = F_{P-CV} \bigotimes \underbrace{\sigma ({MLP}_{CV}(F_{P-E}))}_{CV \, attention} \\
F_{{a.P}-{BEV}} = F_{P-{BEV}} \bigotimes \underbrace{\sigma ({MLP}_{BEV} (F_{P-E}))}_{BEV \, attention}  \\
F_{{a.P}-{RV}} = F_{P-{RV}} \bigotimes \underbrace{\sigma ({MLP}_{RV}(F_{P-E}))}_{RV \, attention}
\end{array}
\right.
\label{eq3}
\end{aligned}
\end{equation}
where $F_{P-CV}$, $F_{P-BEV}$ and $F_{P-RV}$ represent CV, BEV and RV point features, respectively, $F_{a.P-CV}$, $F_{a.P-BEV}$ and $F_{a.P-RV}$ are the corresponding point attention features, $F_{P-E}$ is the extended point features, $\alpha$ is the sigmoid activation function and $\bigotimes$ is the elementwise product operator. After obtaining the attention features, $F_{a.P-CV}$, $F_{a.P-BEV}$ and $F_{a.P-RV}$ are concatenated channel-wise to obtain the fused point features $F_{P-Fusion}$.

We further enrich the fused point features $F_{P-Fusion}$ by concatenating the raw point features $F_{P-Raw}$ from the raw point clouds $P$. To make the raw point features $F_{P-Raw}$ compatible with the attention point features, a simple fully connected network called ${MLP}_{Raw}$ is applied to map the raw point features to appropriate dimensions. The ${MLP}_{Raw}$  is composed of a linear layer, a BN layer, and an ReLU layer. The raw point clouds can partially compensate for the quantization loss of the initial point cloud voxelization.

\begin{figure}[tb]
  \centering
    \includegraphics[scale=0.28]{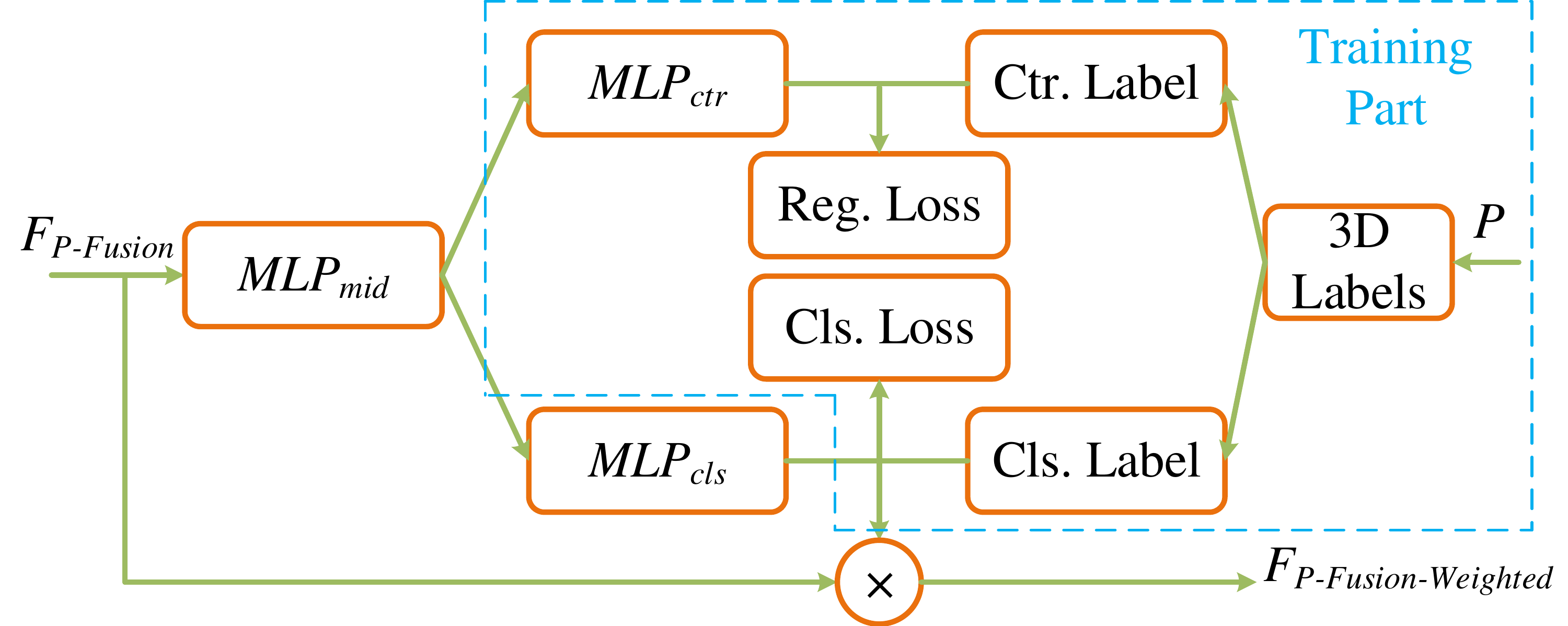}
    \captionsetup{font={footnotesize}}
    \caption{Architecture of the proposed attentive point weighting module.}
    \label{fig5}
\end{figure}

\subsubsection{Attentive Pointwise Weighting}
After the overall scene is encoded by the fused point features $F_{P-Fusion}$, they are grouped into pillars again as inputs of the succeeding fusion feature detection. Since the fused point feature is obtained by interpolation of multi-view 2D feature maps, there is an inevitable loss of 3D geometric and structural information. In addition, most of the points might only represent the background regions. Intuitively, points belonging to the foreground objects should contribute more to final object detection, while those from the background regions should contribute less.
\begin{figure*}[tb]
  \centering
    \includegraphics[scale=0.23]{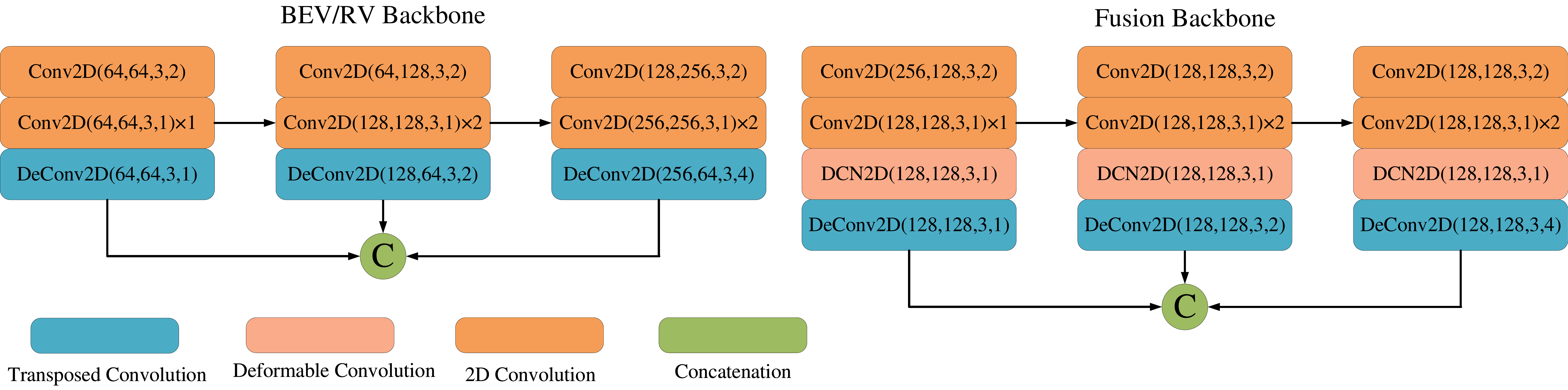}
    \captionsetup{font={footnotesize}}
    \caption{The details of the backbone network for MVAF-Net. Conv(cin, cout, k, s) represents a convolutional block, where cin, cout, k, and s denote the input channel number, output channel number, kernel size and stride, respectively. Each block consists of Convolution, BatchNorm, and Leaky ReLU.}
    \label{fig6}
\end{figure*}
Hence, we propose an attentive pointwise weighting module to perform two extra tasks: foreground classification and pointwise center regression, as shown in Fig. \ref{fig5}. The foreground classification branch is used to predict the foreground/background probability of each point, which is further employed to reweight the fused point features. The center regression branch is used to predict the relative position of each object point to the object center. The center regression can enforce the three-stream backbone networks to learn structure-aware features. The foreground classification and center regression labels can be directly generated by the 3D ground truth bounding boxes, i.e., by checking whether each point is within the 3D bounding boxes and computing the offset from the center of the bounding boxes. The two branches can be formulated as:
%%补充公式
\begin{equation}
\begin{aligned}
\left\lbrace
\begin{array}{l}
F_{cls} ={MLP}_{cls} ({MLP}_{mid} (F_{P-Fusion})) \\
F_{ctr} ={MLP}_{ctr} ({MLP}_{mid} (F_{P-Fusion})) \\
F_{P-Fusion-Weighted} = F_{P-Fusion} \bigotimes F_{cls}
\end{array}
\right.
\label{eq4}
\end{aligned}
\end{equation}
where, ${MLP}_{mid}$, ${MLP}_{cls}$ and ${MLP}_{ctr}$ are fully connected layers, ${MLP}_{mid}$ includes a linear layer and an ReLU layer, ${MLP}_{cls}$ includes a linear layer and a sigmoid layer, while ${MLP}_{ctr}$ only includes a linear layer. $F_{P-Fusion-Weighted}$ is the reweighted and fused point features, $\bigotimes$ is element-wise multiplication operator. The foreground classification loss $L_{fore}$ uses focal loss, the center regression loss $L_{ctr}$ uses SmoothL1 loss, and only the foreground points are considered for center regression loss.

\begin{table*}[tb]
\begin{center}
\footnotesize
\captionsetup{font={footnotesize}}
%\textbf{Table 4}~~Impact of network connection distance.\\
\caption{\textsc{Performance Comparison With Previous Methods on KITTI Test Server. 3D Object Detection Metrics are Used, Reported by the Average Precision With IOU Threshold 0.7. The Bold Values Indicate the Top Performance.}}
\begin{tabular}{p{3cm}<{\centering}p{0.9cm}<{\centering}p{0.9cm}<{\centering}
				p{0.8cm}<{\centering}p{0.8cm}<{\centering}p{0.8cm}<{\centering}p{0.8cm}<{\centering}
				p{0.8cm}<{\centering}p{0.8cm}<{\centering}p{0.8cm}<{\centering}p{0.8cm}<{\centering}
                p{0.5cm}<{\centering}}
\hline
\multicolumn{1}{c}{\multirow {2}{*}{\textbf{Method}}} & \multicolumn{1}{c}{\multirow {2}{*}{\textbf{Modality}}} & \multicolumn{1}{c}{\multirow {2}{*}{\textbf{Stage}}} & \multicolumn{4}{c}{\textbf{3D Detection (Car)}} & \multicolumn{4}{c}{\textbf{BEV Detection (Car)}} & \multicolumn{1}{c}{\multirow {2}{*}{\textbf{FPS}}} \\
\multicolumn{1}{c}{\textbf{}} & \multicolumn{1}{c}{\textbf{}} & \multicolumn{1}{c}{\textbf{}} & \multicolumn{1}{c}{\textbf{Easy}} & \multicolumn{1}{c}{\textbf{Moderate}} & \multicolumn{1}{c}{\textbf{Hard}} &  \multicolumn{1}{c}{\textbf{3D mAP}} & \multicolumn{1}{c}{\textbf{Easy}} & \multicolumn{1}{c}{\textbf{Moderate}} & \multicolumn{1}{c}{\textbf{Hard}} &\multicolumn{1}{c}{\textbf{BEV mAP}} & \multicolumn{1}{c}{\textbf{}}\\
\hline
\hline
PIXOR \cite{yang2018pixor} & L & One & - & - & - & - & 83.97 & 80.01 & 74.31 & 79.43 & - \\
VoxelNet \cite{zhou2018voxelnet} & L& One & 77.82 & 64.17 & 57.51 & 66.5 & 87.95 & 78.39 & 71.29 & 79.21 & 4.4 \\
{SECOND-V1.5} \cite{yan2018second} & L & One & 84.65 & 75.96 & 68.71 & 76.44 & 91.81 & 86.37 & 81.04 & 86.41 & 20 \\
PointPillars \cite{lang2019pointpillars} & L & One & 82.58 & 74.31 & 68.99 & 75.29 & 90.07 & 86.56 & 82.81 & 86.48 & 42 \\
PointRCNN \cite{shi2019pointrcnn} & L & Two & 86.96 & 75.64 & 70.70 & 77.77 & 92.13 & 87.39 & 82.72 & 87.41 & 10 \\
Fast PointRCNN \cite{chen2019fast} & L & Two & 85.29 & 77.40 & 70.24 & 77.64 & 90.87 & 87.84 &80.52 & 86.41 & 15\\
PartA2 \cite{shi2020part} & L & Two & 87.81 & 78.49 & 73.51 & 79.94 & 91.70 & 87.79 & 84.61 & 88.03 & 10\\
MV3D \cite{chen2017multi} & L+C & Two & 74.97 & 63.63 & 54.00 & 64.20 & 86.49 & 79.98 & 72.23 & 78.45 & 2.8 \\
F-PointNet \cite{qi2018frustum} & L+C & Two & 82.19 & 69.79 & 60.59 & 70.86 & 91.17 & 84.67 & 74.77 & 83.54 & 5.9 \\
AVOD-FPN \cite{ku2018joint} & L+C & Two & 83.07 & 71.76 & 65.73 & 73.52 & 90.99 & 84.82 & 79.62 & 85.14 & 10 \\
F-ConvNet \cite{wang2019frustum} & L+C & Two & 87.36 & 76.39 & 66.69 & 76.81 & 91.51 & 85.84 & 76.11 & 84.49 &	2.1 \\
MMF \cite{liang2019multi} & L+C & Two & \textbf{88.40} & 77.43 & 70.22 & 78.68 & 93.67 & 88.21 & 81.99 & 87.96 & 12.5\\
PI-RCNN \cite{xie2020pi} & L+C & Two & 84.37 & 74.82 & 70.03 & 76.41 & 91.44 & 85.81 & 81.00 & 86.08 & 10 \\
EPNet \cite{huang2020epnet} & L+C & Two & 89.81 & \textbf{79.28} & 74.59 & \textbf{81.23} & \textbf{94.22} & \textbf{88.47} & 83.69 & 88.79 & 9 \\
MAFF \cite{zhang2020maff} & L+C & One & 85.52 & 75.04 & 67.61 & 76.06 & 90.79 & 87.34 & 77.66 & 85.26 & \textbf{24} \\
UberATG-ContFuse \cite{liang2018deep} & L+C & One & 83.68 & 68.78 & 61.67 & 71.38 & 94.07 & 85.35 & 75.88 & 85.10 & 16 \\
MVX-Net \cite{sindagi2019mvx} & L+C & One & 85.99 & 75.86 & 70.70 & 77.52 & 91.86 & 86.53 & 81.41 & 86.60 & 6 \\
\bf{Ours} & L+C & One & 87.87 & 78.71 & \textbf{75.48} & 80.69 & 91.95 & 87.73 & \textbf{85.00} & 88.23 & 15 \\
\hline
\hline
\end{tabular}%}
\label{Table1}
\end{center}
\end{table*}
\subsection{Fusion Feature Detection}
In the FFD part, the fused and reweighted point features $F_{P-Fusion-Weighted}$ are again voxelized in the same way as in the BEV stream to obtain pillar features. The encoded pillar features are scattered back to the original pillar positions to construct a pseudo-image. The pseudo-image features are forwarded through fusion backbone for final 3D detection.
We adopt a detection head and loss design similar to\cite{lang2019pointpillars} which comprises three parts: class classification, bounding box regression and direction classification. The class classification loss $L_{cls}$ uses focal loss\cite{loshchilov2018fixing}, bounding box regression loss $L_{loc}$  uses the SmoothL1 function to define the loss of the offset from anchor, and direction classification loss $L_{dir}$ uses a softmax classification loss. The overall loss function can be defined as:
\begin{equation}
L_{total} = \beta_{loc} L_{loc} + \beta_{cls} L_{cls} + \beta_{dir} L_{dir} + \beta_{fore} L_{fore} + \beta_{ctr} L_{ctr}
\end{equation}

\section{Experiment}

In this section, we summarize the dataset in Section A and introduce the implementation details of our proposed MVAF-Net in Section B. In Section C, we evaluate our method on the challenging 3D detection Benchmark KITTI\cite{geiger2013vision}. In Section D, we present ablation studies about our method.

\subsection{Dataset and Evaluation}
We evaluate our proposed MVAF-Net on the KITTI 3D/BEV object detection benchmark. It contains the camera and LiDAR data collected using a single Point Grey camera and Velodyne HDL-64E LiDAR. The training data contains 7,481 annotated frames for RGB images and LiDAR point clouds with 3D bounding boxes for object classes such as cars, pedestrians, and cyclists. Following the common protocol, we further divide the training data into a training set with 3,712 frames and a validation set with 3,769 frames. Additionally, KITTI provides 7,518 frames without labeling for testing. We conduct experiments on the most commonly used car category and use average precision (AP) with an (IOU) threshold of 0.7 as evaluation metrics. The benchmark considers three levels of difficulties: easy, moderate, and hard based on the object size, occlusion state, and truncation level. The average precision (AP) is calculated using 40 recall positions. To further compare the results with other methods on the KITTI 3D detection benchmark, we divide the KITTI training dataset into 4:1 for training and validation and report the performance on the KITTI test dataset.
\begin{figure*}[tb]
    \subfigure[]{
    \includegraphics[scale=0.24]{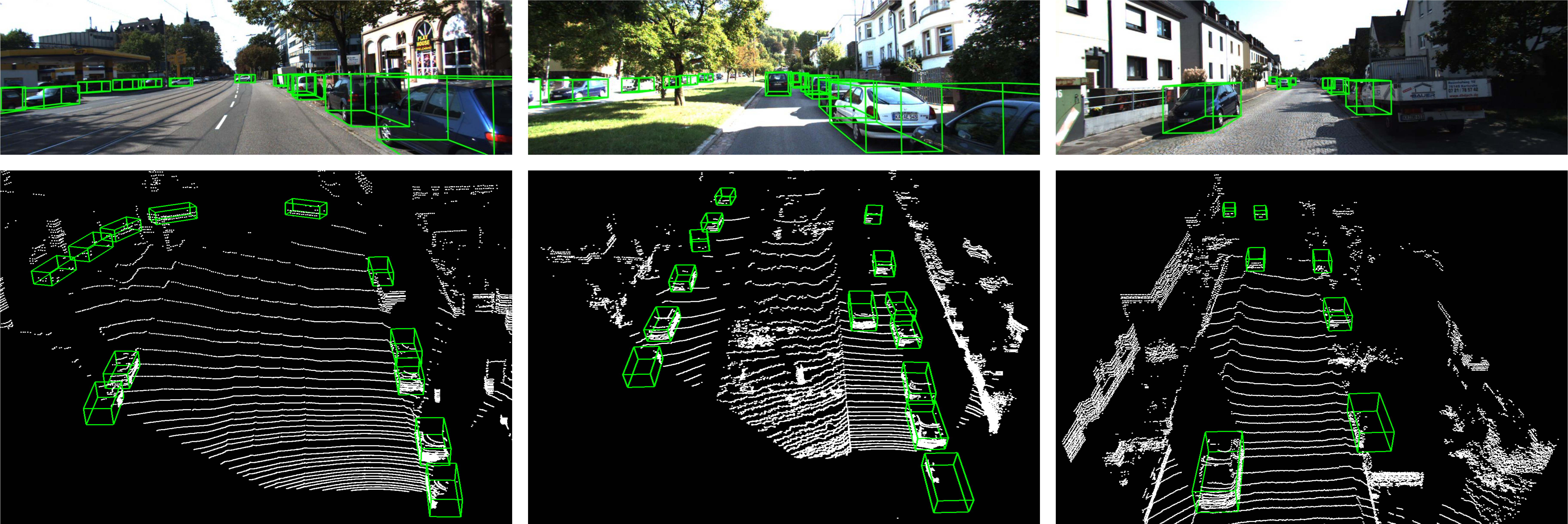}
    }
    \subfigure[]{
    \includegraphics[scale=0.205]{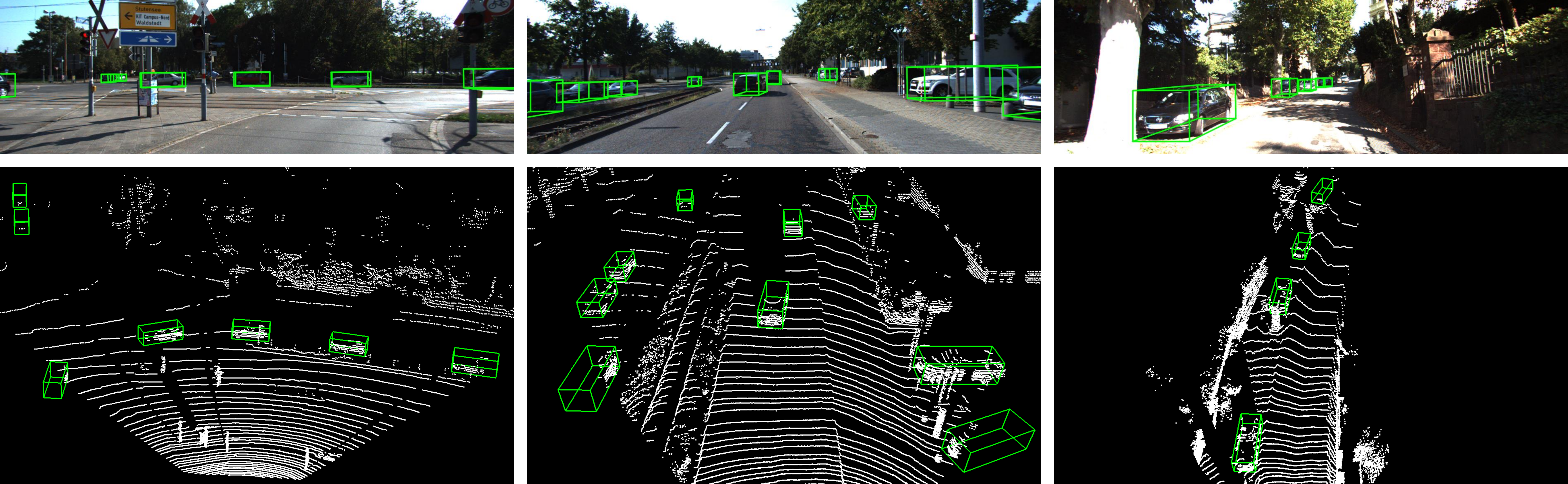}
    }
\captionsetup{font={footnotesize}}
\caption{Qualitative results on the KITTI test set. The predicted bounding boxes are shown in green. The predictions are projected onto the RGB images (upper row) for better visualization.}
\label{fig7}
\end{figure*}

\subsection{Implementation Details}
\subsubsection{Network Settings}
The three-stream backbone takes both the LiDAR point clouds and the RGB images as inputs. For each 3D scene, the range of cropped LiDAR point cloud is [-0,70.4], [-40,40], [-1,3] meters along the $x$, $y$, $z$ axes in LiDAR coordinate, respectively. In the cylindrical coordinate system, the range of the cropped point cloud is [$-\frac{\pi}{2}$, $\frac{\pi}{2}$], [-1, 3] along the $\phi, z$ axes, respectively. A fully connected layer is used to encode the raw point cloud to obtain 64D point features in BEV and RV streams, respectively. We use a voxel size of $v_x=v_y=0.05 \, m$ and $ v_\phi =002454 \, rad$, $v_z=0.05 \, m$ to group the high-dimensional point features into pillars in BEV and RV streams with dynamic voxelization, respectively. Then, a max pooling operation is employed to obtain pillar features that are then scattered back to the original pillar positions to construct a pseudo-image. Thus, we can obtain a regular feature map of size 400$\times$352 and 80$\times$1280 in BEV and RV streams, respectively. The details of our BEV, RV and fusion backbone are shown in Fig. \ref{fig6}, in which the BEV backbone and RV backbone use the same network architecture. To enhance the feature learning of the network, we added a deformable convolution layer to the fusion backbone. The lightweight and efficient RegNetX-200MF is employed as our CV backbone and is initialized with pretrained weights on ImageNet\cite{deng2009imagenet}. For details of the network structure, please refer to \cite{radosavovic2020designing}.

\subsubsection{Training Configuration}
Our entire MVAF-Net network is end to-end trainable, and it is trained by ADAM optimizer with a fixed weight decay of\cite{loshchilov2018fixing} 0.01. The learning schedule is a one-cycle policy with the max learning rate set to 3e-3, the division factor 10, and the momentum range from 0.95 to 0.85. The mini-batch size is set to 2, and the model is trained for 40 epochs. In the detection head, two anchors with different angles $(0^{\circ},90^{\circ})$ were used.

\subsubsection{Data Augmentation}
To guarantee the correct correspondence between LiDAR points and image pixels, we do not use cut-and-paste and individual ground-truth box augmentation strategies when training. This is different from most 3D detection algorithms only based on LiDAR. We only apply random flipping, global rotation, and global scaling to the whole point cloud. The noise for global rotation is uniformly drawn from $[-\frac{\pi}{4}, \frac{\pi}{4}]$, and the scaling factor is uniformly drawn from [0.95,1.05].

\begin{figure*}[tb]
  \centering
    \includegraphics[scale=0.315]{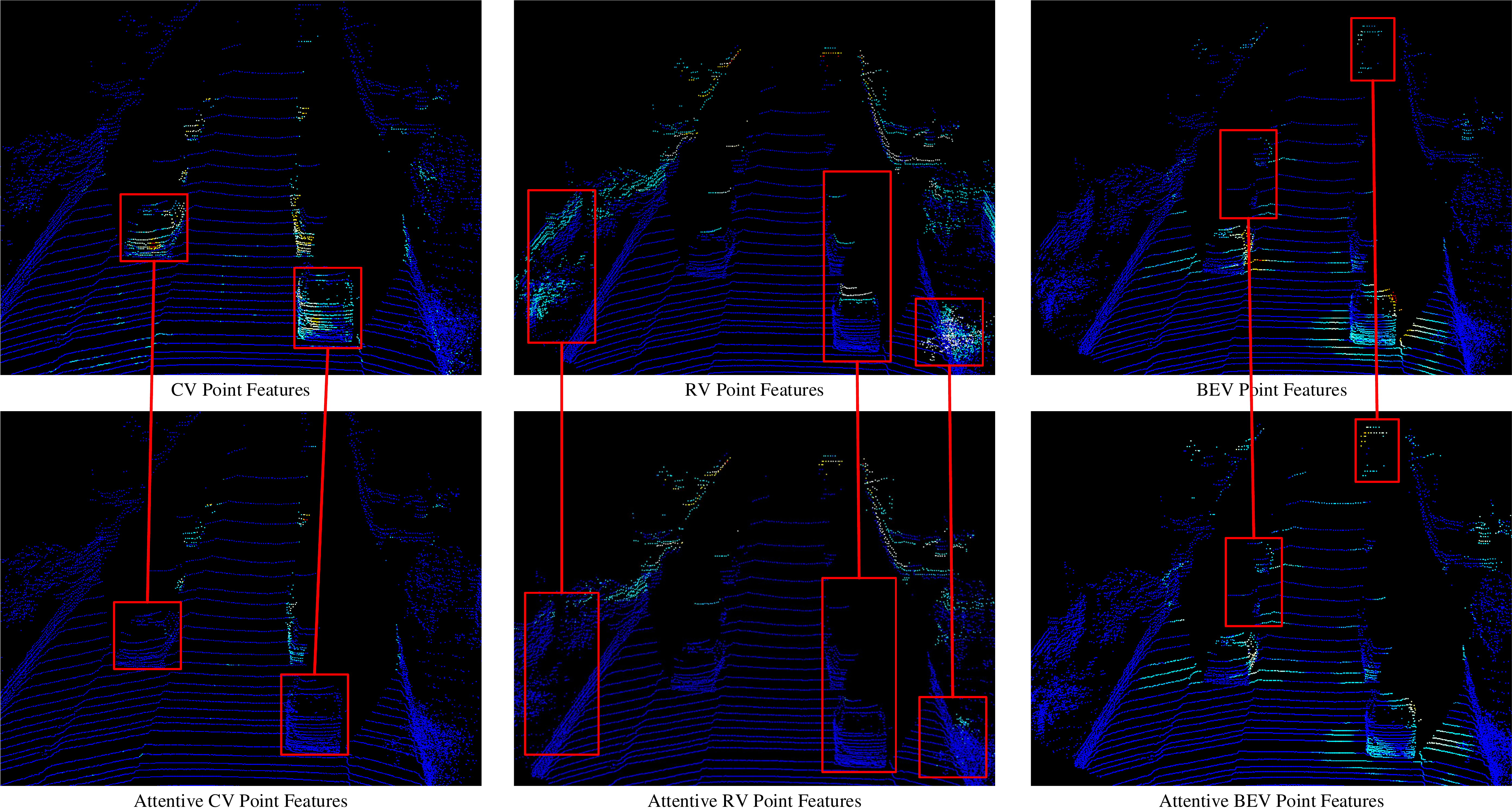}
    \captionsetup{font={footnotesize}}
    \caption{Visualization of point features before/after the attentive pointwise fusion module.}
    \label{fig8}
\end{figure*}

\subsection{Experimental Results on KITTI benchmark}
We compare our single-stage MVAF-Net 3D point cloud detector with other state-of-the-art approaches by submitting the detection results to the KITTI server for evaluation. As shown in Table \ref{Table1}, we evaluate our method on the 3D detection benchmark and the BEV detection benchmark of the KITTI test dataset. It can be observed that the proposed MVAF-Net outperforms multi-sensor based methods  MV3D\cite{chen2017multi}, F-PointNet\cite{qi2018frustum}, AVOD-FPN\cite{ku2018joint}, F-ConvNet\cite{wang2019frustum}, MMF\cite{liang2019multi}, PI-RCNN\cite{xie2020pi}, MAFF\cite{zhang2020maff}, ContFuse\cite{liang2018deep}, and MVX-Net\cite{sindagi2019mvx} by 16.49\%, 9.83\%,7.17\%, 3.88\%, 2.01\%, 4.28\%, 4.63\%, 9.31\% and 3.17\% in terms of the 3D mAP.
In particular, the MVAF-Net achieves the best performance among all single-stage multi-sensor methods and it also achieves competitive results compared with all two-stage multi-sensor methods in both 3D and BEV detection tasks. We can also see that our MVAF-Net achieves the best performance among all methods on the Hard level.
It should be noted that our proposed MVAF-Net is an end-to-end single-stage fusion 3D detection method without extra refinement stage. The MMF\cite{liang2019multi} exploits multiple auxiliary tasks (e.g., 2D detection, ground estimation, and depth completion) to boost the 3D detection performance, which requires many extra annotations. The best multi-sensor method EPNet\cite{huang2020epnet} used the stronger PointNet-based backbone than the 2D Conv backbone that we used, and it adopted a two-stage approach that uses ROI-pooling to fuse the image with the point cloud.
Moreover, the proposed MVAF-Net outperforms all the LiDAR-based 3D object detectors including PartA2\cite{shi2020part}. And, the PartA2\cite{shi2020part} is two-stage method and  also uses the stronger 3D Conv backbone than the 2D Conv backbone used in our MVAF-Net.
These experimental results consistently reveal the superiority of our method over the cascading approach\cite{qi2018frustum,wang2019frustum}, as well as fusion approaches based on ROIs\cite{chen2017multi,ku2018joint} and voxels \cite{liang2019multi,liang2018deep}.

We also visualize several prediction results on the test set in Fig. \ref{fig7}, and we project the 3D bounding boxes from LiDAR coordinates to the RGB images for better visualization. We can see that our approach performs well in capturing cars that are far away, although these objects are recognized with difficulty in RGB images and are susceptible to the sparsity of the point cloud. All these challenging cases persuasively demonstrate the effectiveness of our method.

\subsection{Ablation Studies}
We conduct ablation studies to analyze the effects of the APF and APW module. All models are trained on the training set and evaluated on the validation set of KITTI dataset for Car detection. All evaluations on the validation split are performed via 40 recall positions instead of the 11 recall positions.

\begin{table}[tb]
\begin{center}
\footnotesize
\captionsetup{font={footnotesize}}
\caption{\textsc{Effects of Different Fusion Solutions on KITTI Validation Set For "Car" Detection.}}
\begin{tabular}{p{4cm}<{\centering}ccc}
\hline
\multicolumn{1}{c}{\multirow {2}{*}{\textbf{Method}}} & \multicolumn{3}{c}{\textbf{3D Detection}} \\
\multicolumn{1}{c}{\textbf{}} & \multicolumn{1}{c}{\textbf{Easy}} & \multicolumn{1}{c}{\textbf{Moderate}} & \multicolumn{1}{c}{\textbf{Hard}} \\
\hline
Element-wise Summation & 87.78 & 77.61 & 76.62 \\
Simple Concatenation & 88.54 & 78.71 & 77.48 \\
Attentive Pointwise Fusion & \textbf{89.35} & \textbf{79.56} & \textbf{77.91} \\
\hline
\end{tabular}%}
\label{Table2}
\end{center}
\end{table}

\subsubsection{Effects of Attentive Pointwise Fusion}
We first present comparisons with two alternative fusion solutions: element-wise summation and simple concatenation in Table \ref{Table2}. In element-wise summation (ES)/simple concatenation (SC) fusion method, we use a single fully connected network to map the multi-view point features to the same dimension and element-wise summation/simple concatenation operation is applied to obtain fused point features. As shown in Table \ref{Table2}, our proposed APF module yields an increase in the 3D mAP of $1.57\%/1.95\%/1.29\%$ and $0.81\%/0.85\%/0.43\%$ in Easy/Moderate/Hard level over ES and SC, respectively, which indicates that by considering channel-wise importance in multi-view features, our method can achieve better feature fusion.

To further verify the effectiveness of the proposed APF module, we investigate the importance of each component by removing point features from that view, as shown in Table\ref{Table3}. The 1st row shows that the performance measurably drops if we only aggregate features from BEV, since BEV encoding loses fine-grained information, which is not sufficient for object detection at longer ranges. In the 2nd and 3rd rows, the multi-view features from $F_{P-RV}$, $F_{P-CV}$ contribute significantly to the performance
in all three difficulty levels. The features from $F_{P-RV}$ improve the performance by $3.16\%/4.93\%/6.94\%$, especially in the Hard level, which shows the advantage of RV for long-distance object detection. The features from $F_{P-CV}$ further improve the performance by $0.91\%/0.76\%/0.94\%$. As shown in the last four rows, the additions of raw point features $F_{P-Raw}$ further improve the performance slightly, and the best performance is achieved with all the feature components.

\begin{table}[tb]
\footnotesize
\captionsetup{font={footnotesize}}
\caption{\textsc{Effects of Different Feature Components of APF Module on the KITTI Validation Set for "Car" Detection.}}
\resizebox{0.49\textwidth}{11.5mm}{
\begin{tabular}{ccccccc}
\hline
\multicolumn{1}{c}{\multirow {2}{*}{\textbf{$F_{P-BEV}$}}}  & \multicolumn{1}{c}{\multirow {2}{*}{\textbf{$F_{P-RV}$}}}   & \multicolumn{1}{c}{\multirow {2}{*}{\textbf{$F_{P-CV}$}}}  & \multicolumn{1}{c}{\multirow {2}{*}{\textbf{$F_{P-Raw}$}}}  &  \multicolumn{3}{c}{\textbf{3D Detection}} \\
\multicolumn{1}{c}{\textbf{}} & \multicolumn{1}{c}{\textbf{}} &  \multicolumn{1}{c}{\textbf{}} &  \multicolumn{1}{c}{\textbf{}} &  \multicolumn{1}{c}{\textbf{Easy}} & \multicolumn{1}{c}{\textbf{Moderate}} & \multicolumn{1}{c}{\textbf{Hard}} \\
\hline
 $\surd$ & & & & 84.55 & 73.62 & 69.23\\
 $\surd$ & $\surd$ & & & 87.71 & 78.55 & 76.17 \\
 $\surd$ & $\surd$ & $\surd$ & & 88.62 & 79.31 & 77.11 \\
 $\surd$ & $\surd$ & $\surd$ & $\surd$ & \textbf{89.35} & \textbf{79.56} & \textbf{77.91} \\
\hline
\end{tabular}
}
\label{Table3}
\end{table}
In addition, we also visualize the multi-view point features before APF and after APF, as shown in Fig. \ref{fig8}. We can see that short-distance CV point features are suppressed in the first column, only distant CV point features are used. This shows that the LiDAR point features at short distances are sufficient for the final 3D detection, which is in line with the fact LiDAR points are dense and informative at close distances. In the second column, there are numerous noisy features in the RV point features, especially noisy features from vegetation on both sides of the road at close distances. After the APF module, most of the noise features at close distances are suppressed. Similarly, only the point features of the distant objects are retained, which also indicates that the BEV features at close distances are sufficient for final 3D detection. In the third column, the BEV point features of middle and long distances are enhanced. It can be concluded that our proposed APF module can adaptively estimate the importance of multi-view point features through the attention mechanism to achieve effective multi-view feature fusion.
\begin{figure}[tb]
  \centering
    \includegraphics[scale=0.085]{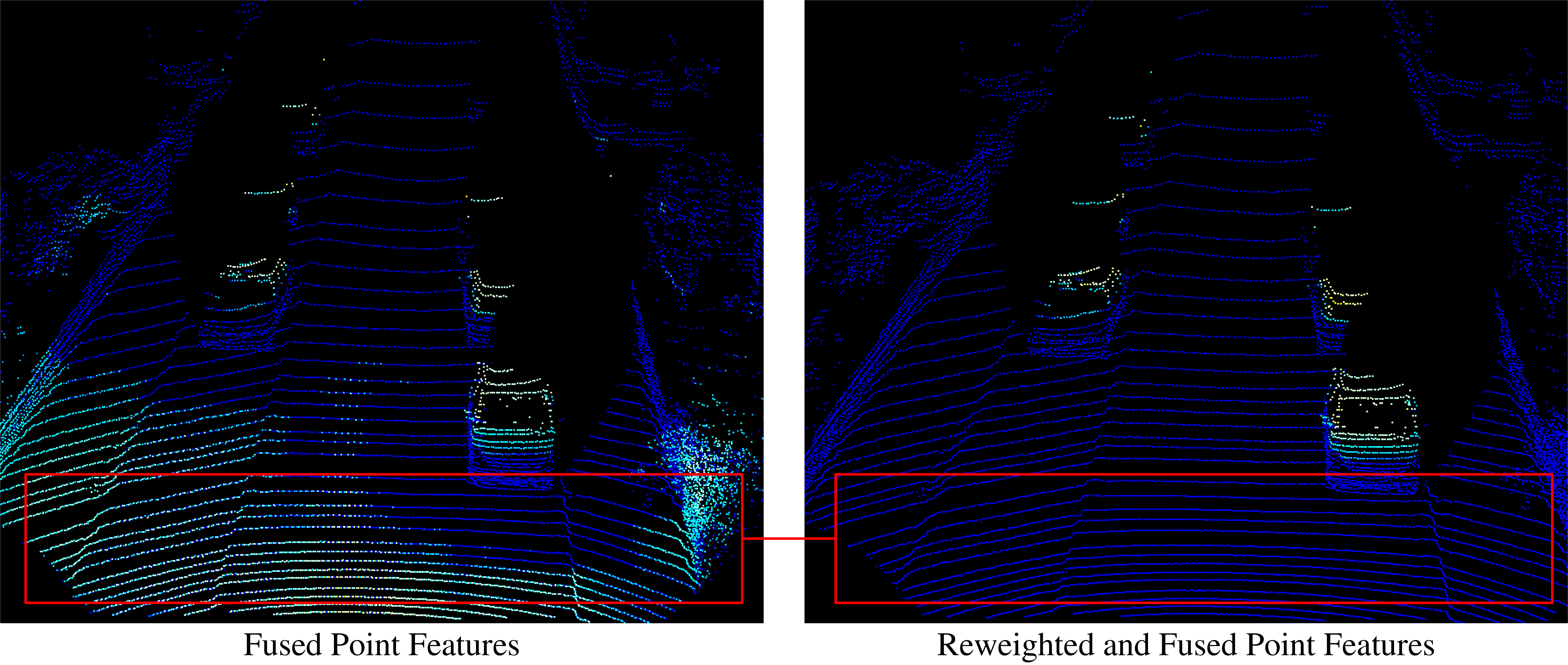}
  \captionsetup{font={footnotesize}}
    \caption{Visualization of point features before/after attentive pointwise weighting module.}
    \label{fig9}
\end{figure}

\begin{table}[tb]
\footnotesize
\captionsetup{font={footnotesize}}
\caption{\textsc{Effects of Different Components of APW Module on the KITTI Validation Set For "Car" Detection.}}
\resizebox{0.49\textwidth}{11.5mm}{
\begin{tabular}{cccccc}
\hline
 \multicolumn{1}{c}{\multirow {2}{*}{\textbf{$F_{cls}$}}}   & \multicolumn{1}{c}{\multirow {2}{*}{\textbf{$F_{ctr}$}}}  & \multicolumn{1}{c}{\multirow {2}{*}{\textbf{$F_{weighted}$}}}  &  \multicolumn{3}{c}{\textbf{3D Detection}} \\
 \multicolumn{1}{c}{\textbf{}} &  \multicolumn{1}{c}{\textbf{}} &  \multicolumn{1}{c}{\textbf{}} &  \multicolumn{1}{c}{\textbf{Easy}} & \multicolumn{1}{c}{\textbf{Moderate}} & \multicolumn{1}{c}{\textbf{Hard}} \\
\hline
 & & & 87.51 & 77.92 & 76.1 \\
 $\surd$ & & & 87.95 & 78.62 & 76.81 \\
$\surd$ & $\surd$ &	& 88.75 & 79.12 & 77.43 \\
 $\surd$ & $\surd$ & $\surd$ & \textbf{89.35} & \textbf{79.56} & \textbf{77.91} \\
\hline
\end{tabular}
}
\label{Table4}
\end{table}
\subsubsection{Effects of Attentive Pointwise Weighting}
We propose the APW module to reweight the point features and learn structure information with extra foreground classification and pointwise center regression. To prove the effectiveness of the proposed APW module, we also investigate the importance of each component of the APW module on the KITTI validation set, as shown in Table \ref{Table4}. The 1st row is the MVAF-Net without the APW module, we can see that the performance drops considerably which shows the effectiveness of the proposed APW module.
The supervisions from $F_{cls}$ and $F_{ctr}$ improve the performance significantly, as shown in the 2nd to 3rd rows. The $F_{cls}$ yields an increase in the 3D mAP of $0.44\%/0.70\%/0.63\%$, the $F_{ctr}$ further improves the performance of $0.80\%/0.50\%/0.62\%$ in Easy/Moderate/Hard level.
As shown in the final 4th row, the use of foreground classification to reweight point features further improves the performance, and the best performance is achieved with all the components. Thus, we can conclude that the proposed APW module enables better feature aggregation by focusing more on the foreground points and learning structure information.

Furthermore, we visualize the fused point features before APW and after APW using point feature reweighting, as shown in Fig. \ref{fig9}. We can see that after the APW module, most of the background point features are suppressed, which shows that our proposed APW is effective.

\section{Conclusion}
We have presented an end-to-end single-stage multi-view fusion 3D detection method, MVAF-Net, which consists of three parts: single view feature Extraction (SVFE), multi- view feature fusion (MVFF) and fusion feature detection (FFD). In the SVFE part, the three-stream CNN backbone (CV, BEV and RV backbone) consumes LiDAR point clouds and RGB images to generate multi-view feature maps. In the MVFF part, the adaptive fusion of multi-view features is realized using our proposed attentive pointwise fusion (APF) module that can adaptively determine how much information is introduced from multi-view inputs using attention mechanisms. Furthermore, we have further improved the performance of the network with the proposed attentive point weighting (APW) module which can reweight the point features and learn structure information with two extra tasks: foreground classification and center regression. Extensive experiments have validated the effectiveness of the proposed APF and APW module. Furthermore, the proposed MVAF-Net yields competitive results and it achieves the best performance among all single-stage fusion methods. Moreover, our MVAF-Net outperforms most two-stage fusion methods, achieving the best trade-off between speed and accuracy on the KITTI benchmark.

\ifCLASSOPTIONcaptionsoff
  \newpage
\fi

\bibliographystyle{IEEEtran}
\bibliography{reference}

\begin{IEEEbiography}[{\includegraphics[width=1.0in,height=1.1in,clip,keepaspectratio]{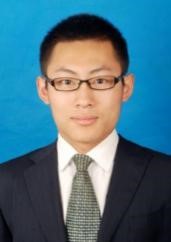}}]{GuoJun Wang}
    received his B.S. degree in vehicle engineering from Yanshan University, Qinhuangdao, China, in 2014. He is currently pursuing his Ph.D. degree in vehicle engineering from Jilin University, Changchun, China. His current research interests include environment perception, behavior estimation and prediction.
\end{IEEEbiography}

\begin{IEEEbiography}[{\includegraphics[width=1.0in,height=1.1in,clip,keepaspectratio]{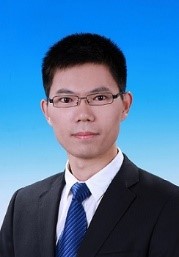}}]{Bin Tian}
   is with the State Key Laboratory of Management and Control for Complex Systems, Institute of Automation, Chinese Academy of Sciences, Beijing 100190, China, and with the Cloud Computing Center, Chinese Academy of Sciences, Dongguan 523808, China (e-mail: bin.tian@ia.ac.cn).
Bin Tian received his B.S. degree from Shandong University, Jinan, China, in 2009 and his Ph.D. degree from the Institute of Automation, Chinese Academy of Sciences, Beijing, China, in 2014. He is currently an Associate Professor of the State Key Laboratory of Management and Control for Complex Systems, Institute of Automation, Chinese Academy of Sciences. His current research interests include computer vision, machine learning, and automated driving.
\end{IEEEbiography}

\begin{IEEEbiography}[{\includegraphics[width=1.0in,height=1.1in,clip,keepaspectratio]{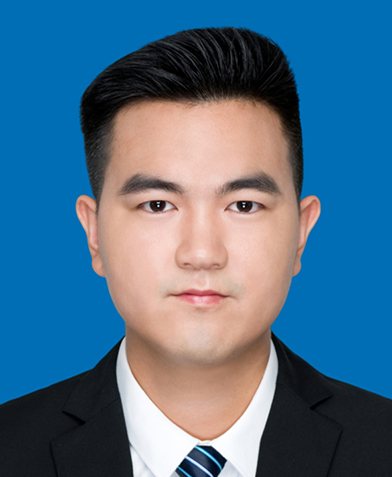}}]{Yachen Zhang}
   received his B.S degree from Sun Yat-Sen University, Guangzhou, China, in 2018. He is currently working toward his M.S. degree in the School of Data and Computer Science, Sun Yat-sen University, China. His research
   interests are in the areas of SLAM and 3D LIDAR. He is now a student under the
   instruction of Long Chen. He is recently focusing on cooperative mapping based on 3D LiDAR.
\end{IEEEbiography}

\begin{IEEEbiography}[{\includegraphics[width=1.0in,height=1.1in,clip,keepaspectratio]{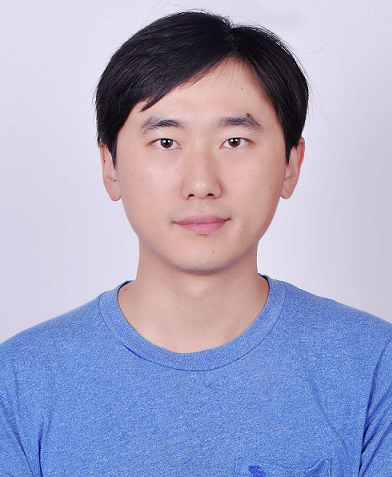}}]{Long Chen}
received his B.Sc. degree in communication engineering and his Ph.D. degree in signal and information processing from Wuhan University, Wuhan, China.He is currently an Associate Professor with the School of Data and Computer Science, Sun Yat-sen University, Guangzhou, China.
%He received the IEEE Vehicular Technology Society 2018 Best Land Transportation Paper Award.
His areas of interest include autonomous driving, robotics, and artificial intelligence, where he has contributed more than 70 publications.
He serves as an Associate Editor for IEEE Transactions on Intelligent Transportation Systems.
\end{IEEEbiography}

\begin{IEEEbiography}[{\includegraphics[width=1.0in,height=1.1in,clip,keepaspectratio]{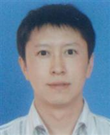}}]{Jian Wu}
 is a Professor of College of Automotive Engineering at Jilin University. His research interests are mainly in vehicle control system, electric vehicles, and intelligent vehicles. He is the authors of over 40 peer-reviewed papers in international journals and conferences, and has been in charge of numerous projects funded by national government and institutional organizations on vehicles.
\end{IEEEbiography}

\begin{IEEEbiography}[{\includegraphics[width=1.0in,height=1.2in,trim = 1mm 0mm 1mm 0mm, clip=true]{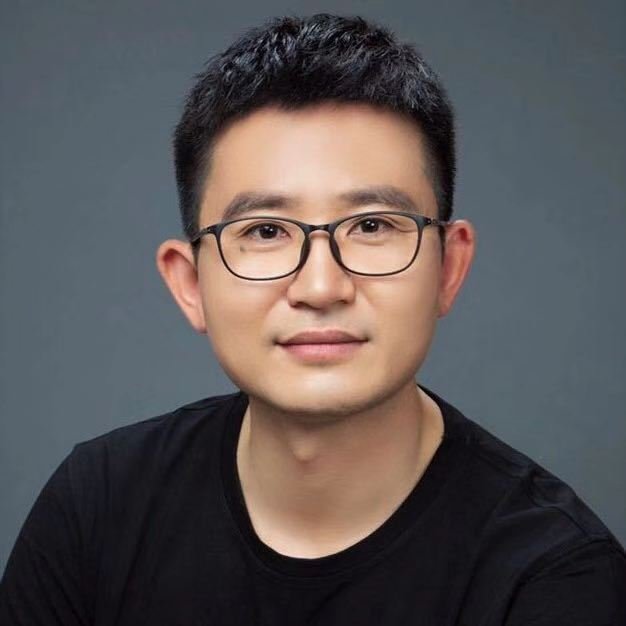}}]{Dongpu Cao}
		received his Ph.D. degree from Concordia University, Canada, in 2008. He is a Canada Research Chair in Driver Cognition and Automated Driving, and currently an Associate Professor and Director of the Waterloo Cognitive Autonomous Driving (CogDrive) Lab at the University of Waterloo, Canada. His current research focuses on driver cognition, automated driving and cognitive autonomous driving. He has contributed more than 180 publications, 2 books and 1 patent. He received the SAE Arch T. Colwell Merit Award in 2012, and three Best Paper Awards from the ASME and IEEE conferences. Dr. Cao serves as an Associate Editor for IEEE TRANSACTIONS ON VEHICULAR TECHNOLOGY, IEEE TRANSACTIONS ON INTELLIGENT TRANSPORTATION SYSTEMS, IEEE/ASME TRANSACTIONS ON MECHATRONICS, IEEE TRANSACTIONS ON INDUSTRIAL ELECTRONICS and ASME JOURNAL OF DYNAMIC SYSTEMS, MEASUREMENT AND CONTROL. He was a Guest Editor for VEHICLE SYSTEM DYNAMICS and IEEE TRANSACTIONS ON SMC: SYSTEMS. He serves on the SAE Vehicle Dynamics Standards Committee and acts as the Co-Chair of the IEEE ITSS Technical Committee on Cooperative Driving.
\end{IEEEbiography}

\end{document}